\documentclass[letterpaper, 10 pt, conference]{ieeeconf}  %

\IEEEoverridecommandlockouts                              %

\usepackage{graphics} %
\usepackage{epsfig} %
\usepackage{mathptmx} %
\usepackage{times} %
\usepackage{amsmath} %
\usepackage{amssymb}  %
\usepackage{subcaption}
\usepackage{booktabs}
\usepackage{url}
\newcommand{\ours}{\textbf{SPARR}}
\title{\LARGE \bf
SPARR: Simulation-based Policies with Asymmetric Real-world Residuals for Assembly
}

\author{Yijie Guo$^{1}$, Iretiayo Akinola$^{1}$, Lars Johannsmeier$^{1}$, Hugo Hadfield$^{1}$, Abhishek Gupta$^{2}$, and Yashraj Narang$^{1}$%
\thanks{$^{1}$Nvidia, $^{2}$University of Washington}%
}

\begin{document}

\setcounter{figure}{1}
\makeatletter
\let\@oldmaketitle\@maketitle%
\renewcommand{\@maketitle}{
   \@oldmaketitle%
   \begin{center}
    \centering      
    \noindent\includegraphics[width=0.95\linewidth]{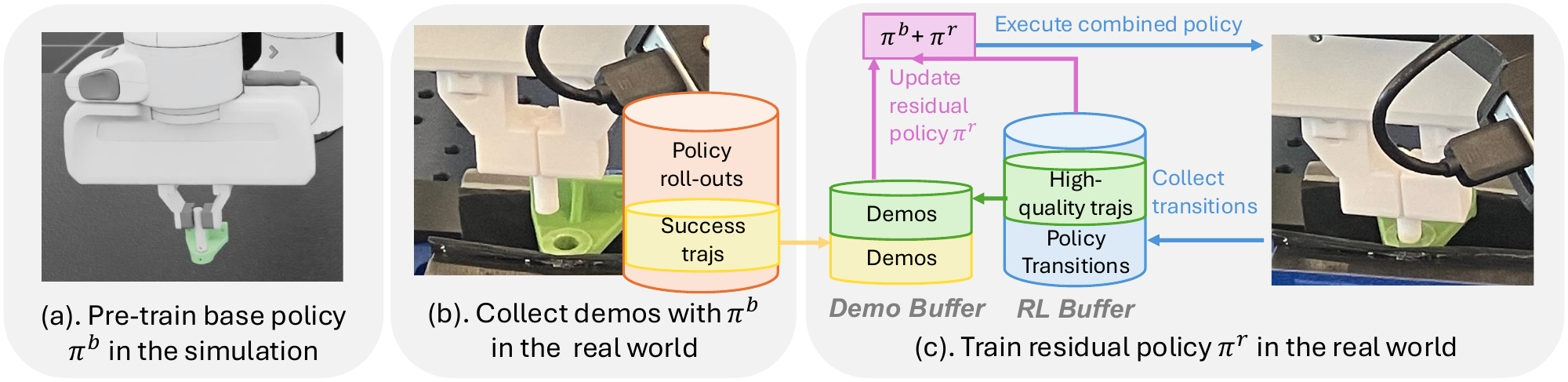}\qquad
  \end{center}
  \vspace*{-0.08in}
  \small{Fig. 1: \textbf{Illustration of our approach, {\ours}.} (a) A specialist policy is pre-trained in simulation. (b) The simulation policy is deployed zero-shot in the real world, achieving a moderate success rate (e.g., up to 80\%). Successful trajectories are collected as demonstrations. (c) A residual policy is trained in the real world on top of the simulation policy, leveraging both the demonstration buffer and the online RL buffer. During training, high-quality trajectories that achieve success quickly are added in demonstrations for further exploitation.}
  \vspace*{-0.17in}

}
\makeatother

\maketitle
\thispagestyle{empty}
\pagestyle{empty}

\begin{abstract}
Robotic assembly presents a long-standing challenge due to its requirement for precise, contact-rich manipulation. While simulation-based learning has enabled the development of robust assembly policies, their performance often degrades when deployed in real-world settings due to the sim-to-real gap. Conversely, real-world reinforcement learning (RL) methods avoid the sim-to-real gap, but rely heavily on human supervision and lack generalization ability to environmental changes. In this work, we propose a hybrid approach that combines a simulation-trained base policy with a real-world residual policy to efficiently adapt to real-world variations. The base policy, trained in simulation using low-level state observations and dense rewards, provides strong priors for initial behavior. The residual policy, learned in the real world using visual observations and sparse rewards, compensates for discrepancies in dynamics and sensor noise. Extensive real-world experiments demonstrate that our method, {\ours}, achieves near-perfect success rates across diverse two-part assembly tasks. Compared to the state-of-the-art zero-shot sim-to-real methods, {\ours} improves success rates by 38.4\% while reducing cycle time by 29.7\%. Moreover, {\ours} requires no human expertise, in contrast to the state-of-the-art real-world RL approaches that depend heavily on human supervision. Please visit the project webpage at \url{https://research.nvidia.com/labs/srl/projects/sparr/}

\end{abstract}

\vspace*{-0.05in}
\section{Introduction}
\vspace*{-0.03in}

Robotic assembly remains a long-standing challenge in robot learning, demanding high-precision, contact-rich manipulation. Simulation and sim-to-real transfer have emerged as powerful strategies for addressing these difficulties. Recent advances in simulation-based learning have led to the development of assembly policies that demonstrate strong performance in both simulated and real world environments \cite{thomas2018learning, fan2019learning, narang2022factory, tian2024asap, tang2023industreal}. Notably, these methods have achieved success rates of up to 80\% on challenging benchmarks of assembly \cite{tang2024automate, tian2025fabrica}.
Despite this progress, current performance is insufficient for industrial deployment, where success rates of 95\% or higher are typically required. Furthermore, simulation-trained policies often exhibit brittleness when deployed directly in the real world due to the sim-to-real gap. For state-based policies, zero-shot performance can degrade significantly due to mismatches in physical parameters (e.g., mass, friction), camera calibration errors, state estimation noise, and variations in grasp pose. Vision-based policies, on the other hand, are especially sensitive to visual domain shifts, such as changes in lighting, object appearance, or background, which can severely impair their real-world generalization.
Meanwhile, recent progress in real-world RL has demonstrated promising results in contact-rich assembly tasks using raw visual observations \cite{luo2024serl, luo2024precise, xu2024rldg}. These methods directly optimize policies in real-world environments, enabling them to capture fine-grained physical interactions. However, they typically rely heavily on human expertise for demonstration collection, as well as active supervision and intervention during training to guide learning.

To address these limitations, we propose {\ours} (\textbf{Fig. 1}), a hybrid framework to pre-train a base policy with low-dimensional state observations in simulation and then learn a residual policy with visual observations in the real world. 
The base policy provides successful demonstrations, a structured prior and safe early exploration, while the residual policy corrects for discrepancies in physical properties, state estimation errors, and visual or environmental differences. This \textit{asymmetric} design enables efficient adaptation to real-world environments without reliance on human supervision.

\begin{figure*}[ht!]
\centering
    \includegraphics[width=0.9\linewidth]{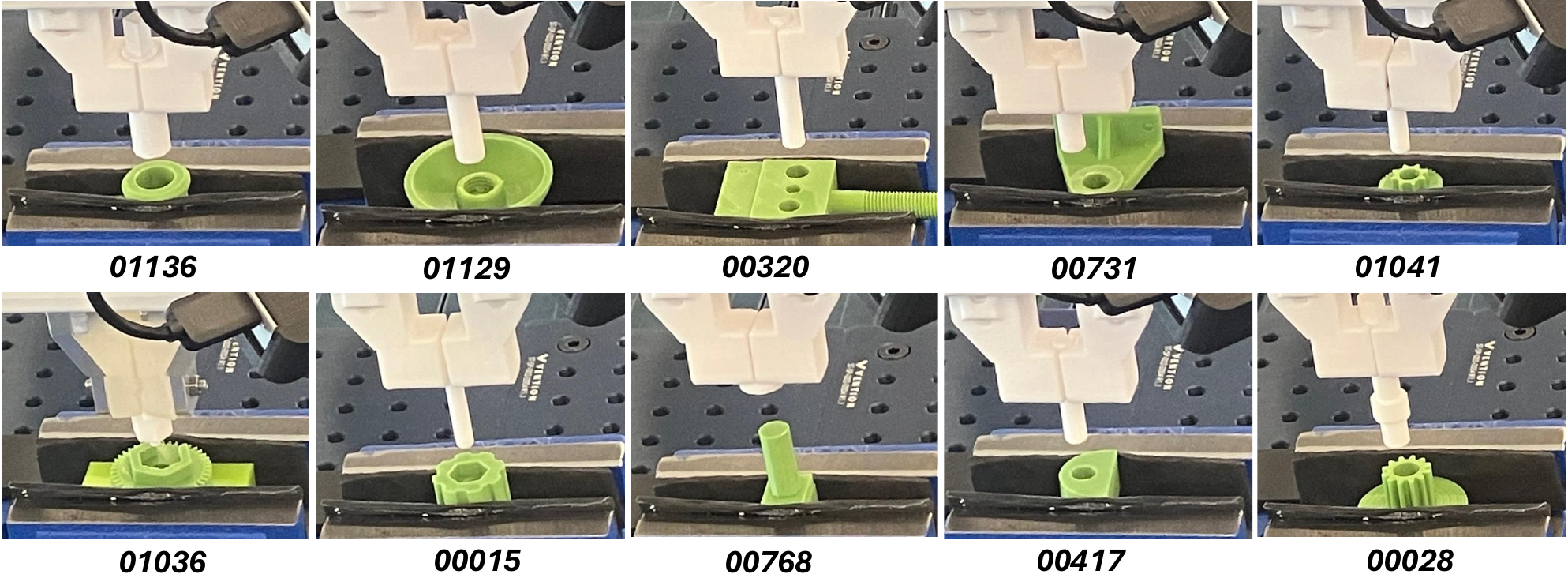}
    \includegraphics[width=0.9\linewidth]{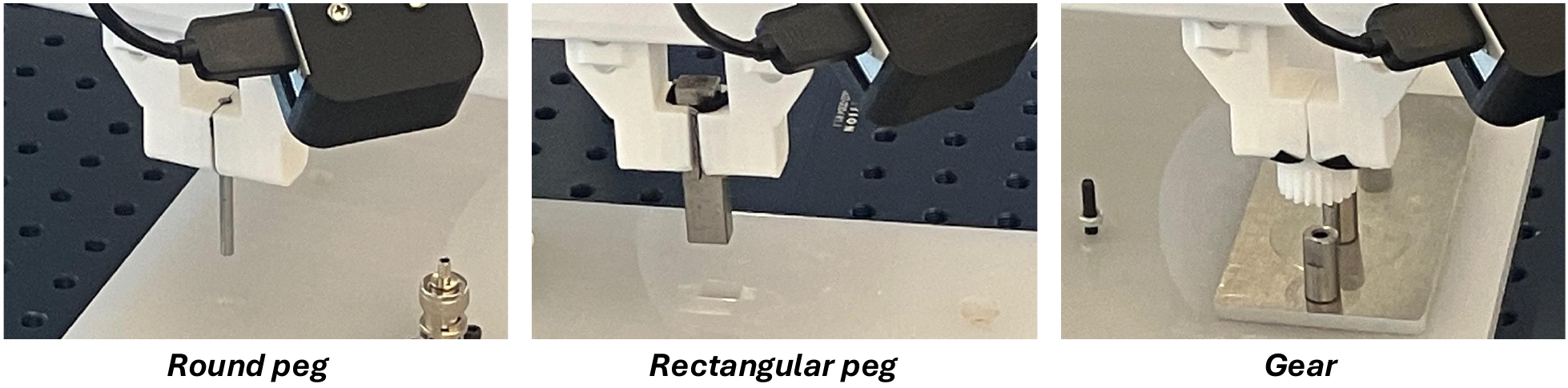}
    \vspace*{-0.08in}
    \caption{\small {\textbf{Overview of experimental tasks}. (Top) 10 AutoMate \cite{tang2024automate} assembly tasks. AutoMate provides a dataset of 100 assembly tasks with diverse parts. We choose 10 out of 100 tasks with near-perfect specialist policy pre-trained in simulation. (Bottom) 3 tasks on NIST board. NIST (National Institute of Standards and Technology) provides Assembly Task Boards as performance benchmarks to evaluate robotic assembly technologies. We consider the peg and gear insertion tasks on task board \#1.  }}
    \label{fig:task}
    \vspace*{-0.2in}
\end{figure*}

In the experiments, {\ours} achieves 95\%-100\% success rates for two-part assembly tasks (Fig.~\ref{fig:task}) in the real-world.
Compared to the state-of-the-art (SOTA) zero-shot deployment \cite{tang2024automate}, {\ours} shows a relative improvement of 38.4\% in success rate and cost 29.7\% less cycle time, averaged over 10 tasks. Unlike the SOTA approach \cite{luo2024precise}, which demands substantial human efforts, {\ours} requires no human involvement in real-world learning.
In summary, our key contributions are:
\begin{itemize}
    \item \textit{Asymmetric Residual Learning Framework}: We introduce {\ours}, a novel framework for sim-to-real transfer, combining a simulation-trained state-based base policy with an asymmetric, vision-conditioned residual policy in the real world.
    This asymmetric design leverages efficient simulation training while enabling robust adaptation to real-world variations.
    \item \textit{Autonomous Adaptation Without Human Supervision}: Unlike existing real-world RL methods that require expert demonstrations or frequent human interventions, {\ours} achieves near-perfect success rates (95–100\%) on real-world assembly tasks with zero human supervision, making it practical for scalable deployment.
    \item \textit{Robustness to State Noise and Physical Variations}: We show that the vision-based residual policy significantly improves robustness to pose estimation errors and socket displacements, outperforming the simulation-trained policy and state-based residual policy.
\end{itemize}

\vspace*{-0.05in}
\section{Related Work}
\vspace*{-0.02in}

\paragraph{Robotic Assembly}
Robotic assembly is a core challenge in manipulation, requiring accurate perception, precise control under contact, and robustness under uncertainty.
Classical approaches based on quasi-static modeling, compliance strategies, and force control have been successful in structured environments \cite{mason1981compliance, whitney1982quasi}, but they typically lack adaptability and scalability to new settings.
Learning-based methods aim to reduce engineering effort and enable generalization. %
Recent work uses deep RL, imitation learning, and skill composition to address the contact-rich nature of assembly \cite{tang2023industreal, tian2024asap, tang2024automate,tian2025fabrica}. Simulation-based learning has made notable progress, 
but performance often degrades in real-world deployment. %
Additionally, recent research on real-world RL directly optimizes assembly in the physical world. \cite{dong2021tactile} leverages tactile sensing but requires human expertise to design curriculum specific for insertion tasks. \cite{luo2024serl} introduces vision-based RL for contact-rich tasks using sparse rewards. %
\cite{luo2024precise} and \cite{xu2024rldg} extend this with human-in-the-loop feedback to achieve sub-millimeter accuracy. %
While these methods push the frontier of real-world robot learning, they often rely on frequent human supervision, intervention, and/or substantial execution time on real-world robots, underscoring the need for more scalable and autonomous solutions.
Our work addresses this issue with a hybrid approach that combines the efficiency and structure of simulation policies with residual adaptation in the real world.

\paragraph{Sim-to-Real Transfer}
Sim-to-real transfer aims to bridge the gap between policies trained in simulation and their deployment in the real world. %
A common strategy is domain randomization, which exposes the policy to randomized physical and visual parameters during training to improve robustness \cite{tobin2017domain, peng2018sim, sadeghi2016cad2rl}.
Especially for image-based policies, this often requires high-fidelity rendering and large-scale training to identify the effective visual augmentations and randomizations \cite{handa2023dextreme}. Yet, such policies can become overly conservative and still fail when the real environment deviates in unmodeled ways.
Alternatively, \cite{bousmalis2018using, james2019sim, chebotar2019closing} explores domain adaptation, where policies or features are adapted post-training to  align with real-world observations.
\cite{zhang2024bridging} fine-tunes simulation policies for tight-insertion tasks with real-world human demonstrations.
Distillation as a widely used adaptation method is time- and compute-intensive \cite{ha2023scaling, yamada2024twist}.
It typically requires significant iterations during real-world development, involving policies rollouts, visual data collection, behavior cloning or DAgger, and fine-tuning.

\paragraph{Residual Policy Learning for Assembly Tasks}
Residual policy learning has emerged as an effective strategy to enhancce base policies with corrective behaviors, particularly in contact-rich manipulation.
Early work \cite{johannink2018residual, johannink2019residual, kulkarni2022learning} augmentes conventional feedback controllers with RL-based residuals to handle unmodeled contact dynamics, demonstrating success in large-part assemblies.
\cite{zhang2023efficient} learns compliance parameter adjustments via real-time residuals atop a simulation policy. This work supports assembly, pivoting, and screwing tasks though limited to large-part assemblies (with edge lengths exceeding 4 cm).
Recently, \cite{ankile2024imitation} combines behavior cloning with a residual RL policy, achieving sim-to-real transfer for high-precision assembly, but it requires training both the base and residual policies in simulation as well as collecting human demonstrations.
\cite{zhang2024residual} explores learning a vision-based base policy and a force-based residual policy, but primarily evaluates assembly tasks in simulation.
\cite{xu2025compliant} proposes compliant residual DAgger with force feedback and compliance control, but relies on human corrections.
Across these works, the residual learning paradigm enables compact powerful adaptations but often depend on human expertise for controller design or demonstration collection.
In contrast, our approach leverages simulation as a platform to alleviate the need for human demonstrations and is applied to fine-grained assembly tasks (average diameter of plugs $<1cm$). %

\vspace*{-0.05in}
\section{Method}
\vspace*{-0.03in}
\label{sec:method}

\subsection{Problem Description}
\vspace*{-0.03in}

This work studies policy adaptation from simulation to the real world for two-part assembly tasks (Fig.~\ref{fig:task}). Each environment consists of a Franka robot \cite{haddadin2024franka}, a \textit{plug} (the part to be inserted), and a \textit{socket} (the corresponding receptacle). The objective is to insert the plug into its designated socket.
We formulate the task as a reinforcement learning (RL) problem under the standard Markov decision process (MDP) framework, $M=\{S, A, \rho, P, r, \gamma\}$.
At each time step $t$, $s_t\in S$ is
the state observation (e.g., the robot’s proprioceptive signals and part poses or visual observations),
$a_t \in A$ is the action (e.g., delta end-effector pose), and $r_t$ is the reward. The initial state $s_0$ is drawn from the distribution $\rho(s_0)$ and transitions follow unknown, possibly stochastic dynamics $P$. The agent seeks a policy $\pi(a_t | s_t)$ that maximizes the accumulative discounted rewards, and $\gamma$ is the discount factor.

Our approach is motivated by the practical deployment of robotic assembly policies in industrial and manufacturing settings. A simulation-trained policy typically transfers to the real world with a moderate success rate (e.g., up to 80\% in \cite{tang2023industreal, tang2024automate, tian2025fabrica}).
By training a residual policy on a given set of plug-socket pairs in the real world, we aim to boost performance to near-perfect levels. Once trained, the residual policy is combined with the simulation-based base policy and deployed to assemble new plug-socket pairs directly on the assembly line. 
An overview is shown in Fig. 1.
In the following section, we describe base policy pre-training in simulation (Sec.~\ref{sec:pre_training}), demonstration collection (Sec.~\ref{sec:demo}), and residual policy learning in the real world (Sec.~\ref{sec:training}).

\vspace*{-0.02in}
\subsection{Pre-training in Simulation}
\vspace*{-0.03in}
\label{sec:pre_training}
As depicted in Fig. 1(a) and Fig.~\ref{fig:combined}, we train a base policy $\pi^b$ using low-dimensional state-based observations.
State-based training in simulation is computationally efficient and robust to variations in object poses and visual patterns when transferred to the real world. In contrast, image-based policies require extensive domain randomization and large-scale training~\cite{handa2023dextreme} to achieve robustness against changes in dynamics, object poses, and visual domain gaps.

Following AutoMate \cite{tang2024automate}, the leading baseline for zero-shot sim-to-real transfer in assembly, our state observations $s^b_t$ includes robot joint angles, current end-effector pose, goal end-effector pose, and their difference.
The action $a_t$ is defined as an incremental pose target, which is tracked by a Cartesian impedance controller.
We train the policy $\pi^b$ with Proximal Policy Optimization (PPO) \cite{schulman2017proximal} using imitation rewards derived from disassembly trajectories. $\pi^b$ outputs Gaussian-distributed actions $a_t\sim N(\mu_t, \sigma_t)$ for continuous control.
To enhance robustness, we randomize robot joint configurations, socket poses, and plug poses at the start of each episode.
Thanks to reliable state information, dense reward signals, and parallelized simulation environments, the specialist policy can be trained effectively and efficiently to achieve a high success rate in simulation with randomized initial states, even exceeding 99\% in some tasks \cite{tang2024automate}.

\vspace*{-0.03in}
\subsection{Policy Formulation in the Real World}
\vspace*{-0.03in}
\label{sec:input}

\paragraph{Deployment of Base Policy} When deployed in the real world, $\pi^b$ takes the end-effector goal pose as part of $s^b_t$.
We obtain the socket pose via a \textit{pose-estimation pipeline} combining Grounding DINO \cite{liu2024grounding}, SAM2 \cite{ravi2024sam}, and FoundationPose \cite{wen2024foundationpose}.
Then, we set the end-effector goal pose as the estimated socket pose with a z-axis offset, assuming the plug is consistently held in the gripper.
For comparison, we also obtain ground-truth goal poses by manually guiding the Franka arm to complete the insertion.
On AutoMate assemblies, the difference between our estimated and ground-truth poses is within $\pm1mm$ along the $x$ and $y$ axes.
Based on this observation, we \textit{model state-estimation noise} by uniformly sampling up to $1mm$ in the $x$ and $y$ axes
and adding it to the ground-truth goal pose at the start of each episode. We deliberately avoid directly using estimated poses as inputs, which could cause overfitting to the error distribution from our specific pose estimator. For initialization in each episode, the end-effector holding the plug is positioned $2cm$ above the noisy goal pose.
In the real world, $\pi^b$ achieves only a moderate zero-shot success rate due to dynamics mismatch and state-input inaccuracies. We therefore introduce a residual policy $\pi^r$ to adapt $\pi^b$ to real-world environments that may differ from pre-training.

\begin{figure}[t!]
\centering
    \includegraphics[width=0.95\linewidth]{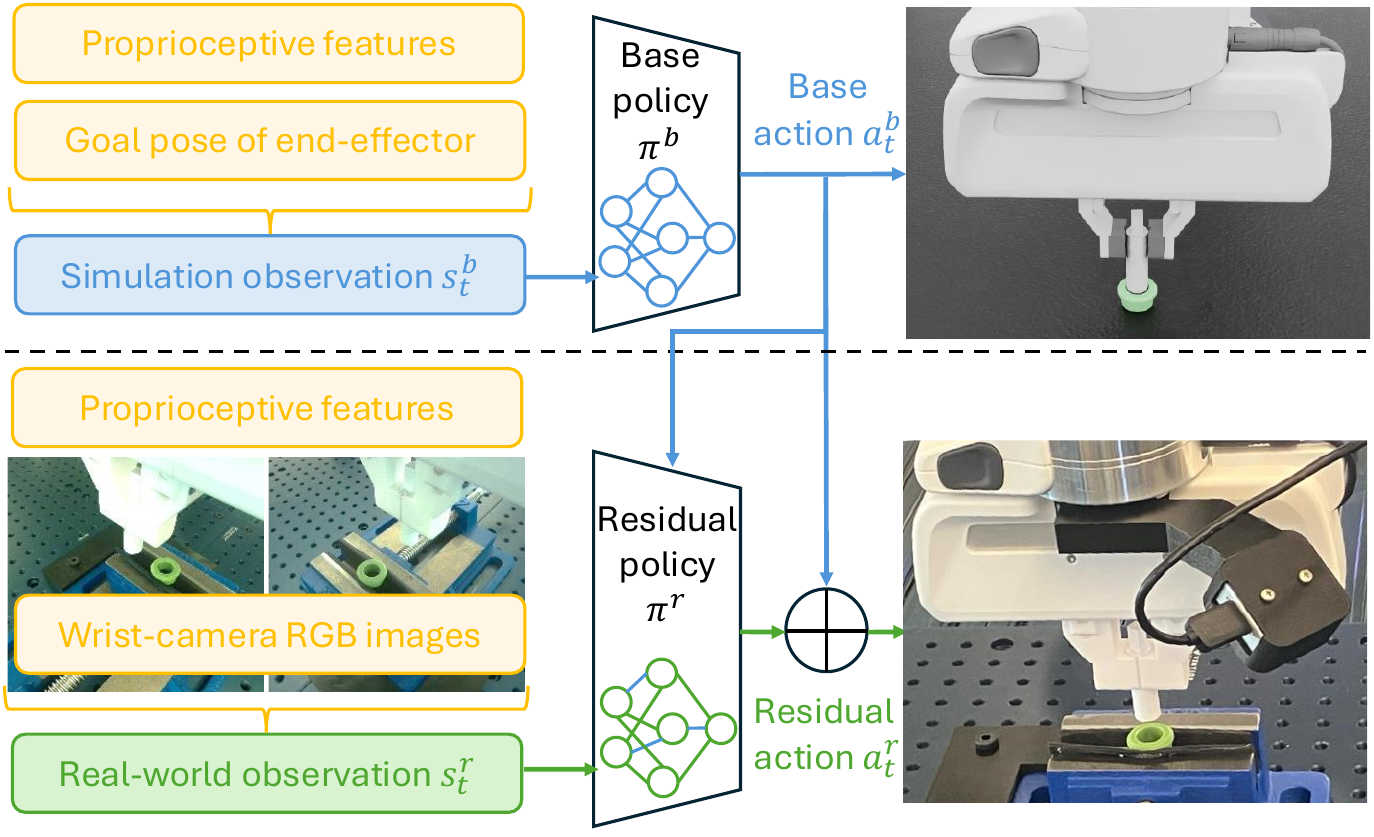}
    \vspace*{-0.1in}
    \caption{\small{\textbf{Illustration of asymmetric policy combination}. Combining state-based base policies from simulation and image-based residual policies learned in the real-world.}}
    \label{fig:combined}
    \vspace*{-0.2in}
\end{figure}

\paragraph{Design of Residual Policy}
As shown in Fig.~\ref{fig:combined}, $\pi^r$ takes as input $s^r_t$, which includes both low-dimensional state (end-effector pose, linear and angular velocities, and force and torques \cite{luo2024precise}) and visual observations (two RGB images). These inputs \textit{exclude any object pose} and are not affected by state-estimation errors.
In addition, visual inputs provide complementary information not captured in the simulated CAD model, such as object asymmetries, textures, manufacturing tolerances and defects, and fine-grained contact features. A CAD-based pose estimator cannot capture geometric constraints such as a USB socket’s one-way insertion, whereas visual observations can.

The residual policy $\pi^r$ outputs actions $a^r_t$ as incremental pose targets, added to $a^b_t$ from $\pi^b$.
Unlike in simulation, dense reward signals (e.g., disassembly path imitation \cite{tang2024automate} or signed distance field metrics \cite{tang2023industreal}) are not available in the real world.
Also, typical dense-reward terms, such as the negative distance to the goal, can be misleading, since surrounding geometry (e.g., socket walls) may obstruct the plug.
We therefore use a sparse success reward: the task succeeds if the current end-effector pose is within $3mm$ in translation and $5^{\circ}$ in rotation of the goal, yielding $r_t = 1$.

\vspace*{-0.03in}
\subsection{Demonstration Collection in the Real World}
\vspace*{-0.03in}
\label{sec:demo}
Instead of training the residual policy $\pi^r$ from scratch in the real world, we propose using rollouts of the simulation policy $\pi^b$ to collect demonstrations for the residual. Since ground-truth residual actions are not available, we generate pseudo-labels by sampling residual actions as Gaussian noise based on the action distribution from $\pi^b$.
At each timestep $t$, we gather proprioceptive features with the noisy goal pose as $s^b_t$ and capture camera images for observation $s^r_t$. The base policy $\pi^b$ is queried with $s^b_t$ to produce a Gaussian distribution of actions $N(\mu_t, \sigma_t)$.
We use the mean $\mu_t$ as the deterministic, base action, i.e. $a^b_t = \mu_t$. 
The residual action $a^r_t$ is then sampled from $N(0, \sigma_t)$. 
The combined action is defined as $a_t=a^b_t+a^r_t$, which ensures that $a_t$ follows the same distribution $N(\mu_t, \sigma_t)$ as the base policy output.

We record the resulting trajectory as a sequence of transitions $\tau=\{(s^r_0, a^b_0, a^r_0, r_0), \cdots, (s^r_t, a^b_t, a^r_t, r_t), \cdots \}$. 
As in Fig. 1(b), zero-shot deployment of $\pi^b$ with injected residual actions yields a subset of successful trajectories, which we store as demonstrations to bootstrap training of $\pi^r$.

\vspace*{-0.03in}
\subsection{Residual Policy Learning in the Real World}
\vspace*{-0.03in}
\label{sec:training}

We propose learning a policy $\pi^r$ to predict residual actions $a^r_t \sim \pi^r(\cdot | s^r_t, a^b_t)$, conditioned on both real-world observations and the base action (Fig.~\ref{fig:combined}). 
As explained in Sec.~\ref{sec:input}, real-world observations $s^r_t$ provide information that is robust to state noise.
Conditioning on the base action $a^b_t$ further supplies context for determining the appropriate correction. We empirically evaluate this design choice in Sec.~\ref{sec:exp_ablation}.

Learning $\pi^r$ in the real world is challenging due to sparse rewards and limited interaction data. To address these challenges, we adopt the RLPD algorithm~\cite{ball2023efficient}, which is sample efficient and capable of leveraging prior demonstration data.
As shown in Fig. 1(c), at each policy update, transitions are sampled equally from a demonstration buffer and an RL buffer. The demonstration buffer is initialized with the offline success trajectories collected in Sec.~\ref{sec:demo}. We then continue to update the demonstration buffer with high-quality trajectories encountered during residual policy learning \cite{oh2018self}. Specifically, trajectories that achieve task success in fewer time steps than the median of prior demonstrations are added to the buffer, allowing the policy to take advantage of strong experiences gathered during training and decrease cycle time. We further analyze this design choice in Sec.~\ref{sec:exp_ablation}.

\begin{figure*}[ht!]
\centering
    \includegraphics[width=0.95\linewidth]{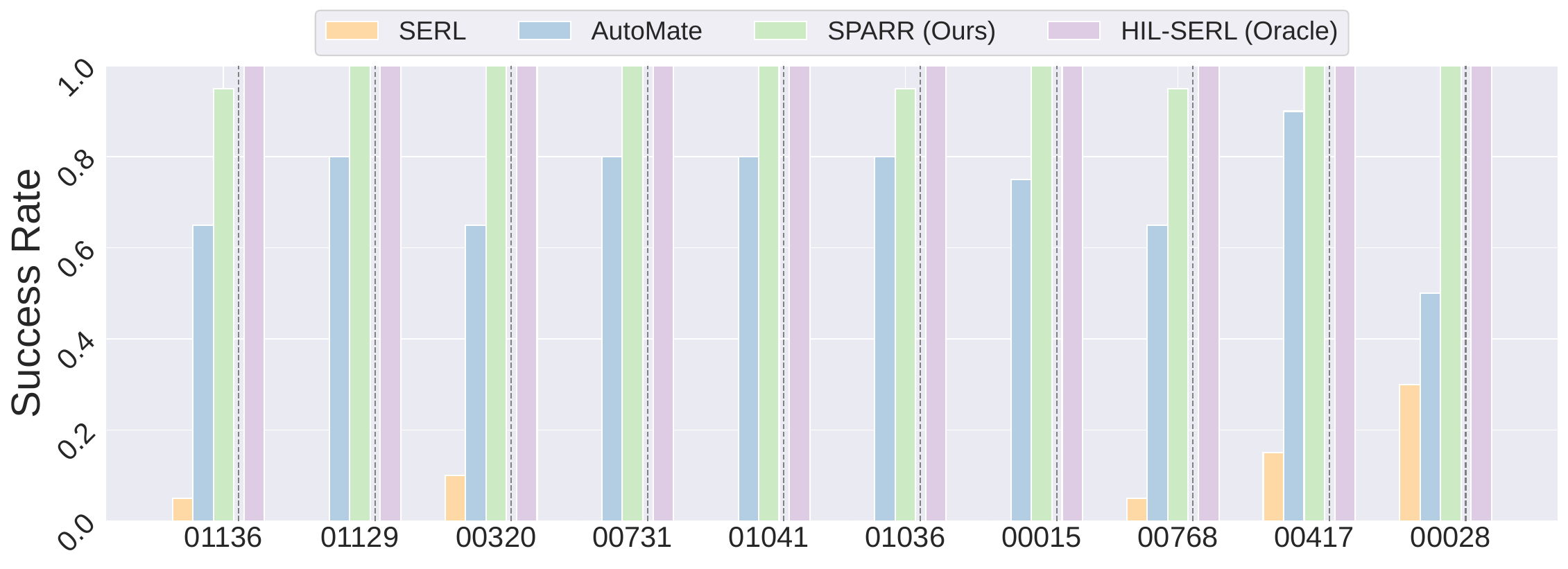}
    \includegraphics[width=0.95\linewidth]{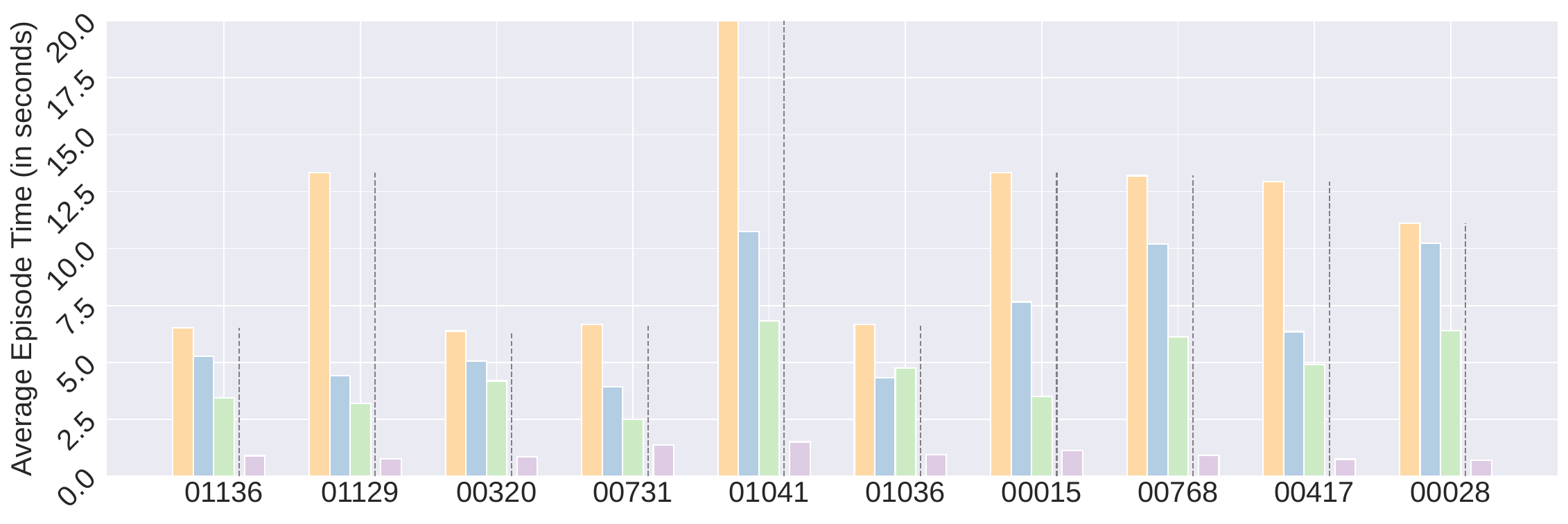}
    \vspace*{-0.08in}
    \caption{\small{\textbf{Performance on 10 AutoMate tasks}. We evaluate the success rate ($\uparrow$ higher is better) and cycle time ($\downarrow$ lower is better) averaged over 20 episodes. SERL, AutoMate, and {\ours} (Ours) transfer simulation-trained policies to the real world without human effort, where \textbf{{\ours}} achieves substantially higher success rates and shorter cycle times. HIL-SERL (Oracle) serves as an upper bound, assuming access to near-optimal human demonstrations and continuous human supervision.}}
    \label{fig:automate_perf}
    \vspace*{-0.15in}
\end{figure*}

\section{Experiments}
In this section, we design experiments to answer the following questions: (1) Can {\ours} effectively and efficiently adapt simulation policies to real-world assembly tasks with near-perfect success rates, without human demonstrations or interventions? (2) Can {\ours} achieve robustness to pose variations and pose-estimation errors? (3) Can {\ours} adapt simulation policies to unseen tasks in the real world?

\vspace*{-0.03in}
\subsection{Setup}
\vspace*{-0.03in}
In our experiments we use the following components:
\begin{itemize}
    \item A Franka Emika robot \cite{haddadin2024franka}, 
    \item an NVIDIA Jetson Orin GPU to run robot controllers,
    \item two Intel RealSense D405 cameras rigidly connected to the flange of the robot,
    \item a PC with an NVIDIA RTX 4090 GPU to run policies,
    \item a SpaceMouse input device only used for \cite{luo2024precise}.
\end{itemize}

\vspace*{-0.03in}
\subsection{Effective Real-world Adaptation}
\vspace*{-0.03in}
\label{sec:exp_automate}
We investigate 10 real-world robotic assembly tasks from the AutoMate dataset~\cite{tang2024automate}. We pre-train base policies in Isaac Lab~\cite{mittal2023orbit} using 128 parallel environments, completing 25 million environment steps within one day of training. 
We select 10 out of 100 tasks (Fig.~\ref{fig:task}) that achieve over 99\% success in simulation. We choose them for their diverse geometries and strong simulation performance, which are expected to perform well in the real world. 
We compare following approaches that deploy simulation policies in the real world without human demonstrations:
\begin{itemize}
    \item \textbf{SERL: a real-world RL approach for precise, contact-rich tasks} \cite{luo2024serl}. Rather than using human demonstrations, we roll out the simulation policy, collect 20 successful trajectories and then run SERL for 0.5 hours to learn real-world assembly policies. %
    \item \textbf{AutoMate: the SOTA approach of zero-shot sim-to-real transfer for two-part assembly} \cite{tang2024automate}. We deploy the simulation policy directly in a zero-shot manner. 
    \item  \textbf{{\ours} (Ours)}. We collect 20 successful trajectories with the base policy from simulation but then trains a residual policy on top of the base policy for 0.5 hours.
\end{itemize}

\begin{figure*}[t!]
\centering
  \includegraphics[width=0.95\linewidth]{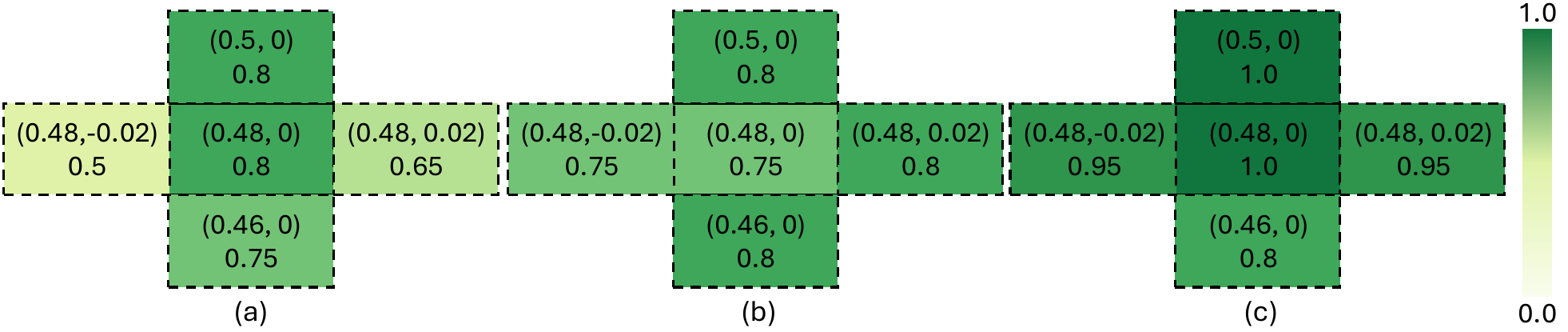}
    \vspace*{-0.1in}
    \caption{\small{\textbf{Policy deployment on different socket poses for AutoMate task 00731.} Each box indicates the (x, y) coordinates of the socket pose and the corresponding success rate (0–1) during evaluation. The training socket pose is at (0.48, 0), and for evaluation, the socket is displaced by 2$cm$: up (0.50, 0), down (0.46, 0), left (0.48, –0.02), and right (0.48, 0.02). (a) Base policy from simulation. (b) Base policy with a state-based residual policy. (c) {\ours} (Ours): base policy with image-based residual policy. The color bar represents success rate from 0 (yellow) to 1 (green). {\ours} achieves higher success rates (darker green) and demonstrates robustness to socket pose variations.}}
    \label{fig:state_noise}
    \vspace*{-0.15in}
\end{figure*}

Additionally, to establish an upper bound on performance, we run \textbf{HIL-SERL, the SOTA real-world RL approach} \cite{luo2024precise} where human experts provide near-optimal demonstrations, frequent interventions, and consistent supervision during training. We manually collect 20 demonstrations with a space mouse and train aseembly policies for 0.5 hours with frequent human interventions to ensure nearly all episodes succeed during real-world training.

We fix the training budget to 0.5 hours across all methods, as real-robot training is expensive and shorter training emphasizes data efficiency. This choice also aligns with recent real-world adaptation and fine-tuning works, which demonstrate effective learning within 30 minutes or less~\cite{hu2025robot, kulkarni2022learning}. All actions are executed at 15 Hz. We evaluate policies over 20 episodes with noisy goal poses (Sec.~\ref{sec:input}).

In Fig.~\ref{fig:automate_perf}, \textbf{SERL}  
exhibits poor performance due to hard exploration in the sparse-reward setting. With only 20 demonstrations and 0.5 hours of real-world training, it struggles to discover reasonable behaviors and collect positive rewards, as no human intervention is provided during online training.
While a few successful trials are observed during real-world learning, SERL cannot efficiently learn to reproduce these successes.
\textbf{AutoMate} shows a moderate success rate despite strong simulation performance, indicating the sim-to-real gap.
In comparison to AutoMate, \textbf{\ours} achieves a relative improvement of 38.4\% in success rate and 29.7\% in cycle time, highlighting the effectiveness of the residual policy in correcting actions of the simulation policy. Notably, {\ours} attains a 95–100\% success rate, comparable to HIL-SERL, without any human supervision or interventions.
\textbf{HIL-SERL}, as an oracle approach, shows reduced cycle times compared to other methods that lack human demonstrations or interventions. We attribute this to two main factors: First, sim-to-real transfer approaches use action smoothing and a policy-level action integrator (PLAI) \cite{tang2023industreal} for reliable deployment in the real world, which can negatively affect policy execution speed.
Second, the quality of demonstration data differs: faster human demonstrations lead to faster learned policies. Since we constrain real-world training to 0.5 hours per task, RL cannot sufficiently overcome the prior from demonstrations. We leave it to future work to reduce cycle time without increasing training time or human efforts.

\vspace*{-0.05in}
\subsection{Robustness to State Noise}
\vspace*{-0.03in}

In high-mix, low-volume manufacturing, robustness to part pose variations is a compelling capability.
In particular, the socket pose on an assembly line may not exactly match the pose set during residual policy learning. To assess performance under such conditions, we physically displace the socket by $2cm$ from its training position. We also add a uniform noise of $1mm$ to the ground-truth goal pose at each episode, to emulate pose-estimation error, as explained in Sec.~\ref{sec:input}.
Here two sources of noise are considered: (1) \textit{part pose variation}, due to absence of precise fixtures in flexible automation, and (2) \textit{pose estimation error} since the exact goal pose is unavailable at deployment.

We compare {\ours} (Ours) with a variant using a state-based residual policy. Our image-based residual policy takes as input two RGB images from wrist cameras, along with the end-effector pose, velocity, force, and torques. The state-based alternative uses the same proprioceptive inputs plus the noisy goal pose instead of images. 
In Fig.~\ref{fig:state_noise}, the \textit{state-based residual policy} is sensitive to errors in estimated goal pose and to out-of-distribution socket positions, showing degraded performance.
In contrast, our \textit{image-based residual policy} is less affected by socket pose changes, as long as the visual observation remains similar to the training distribution. Its performance only drops slightly under extreme pose shifts, primarily due to the resulting changes in the base action distribution.
At novel socket poses, the base policy may produce out-of-distribution base actions that the residual policy cannot fully correct. Overall, {\ours} with the image-based residual policy achieves a 38.6\% improvement over the base policy and outperforms the state-based residual policy by 20.8\%, averaged across varying physical socket positions.

\begin{figure}[ht!]
\centering
  \includegraphics[width=0.95\linewidth]{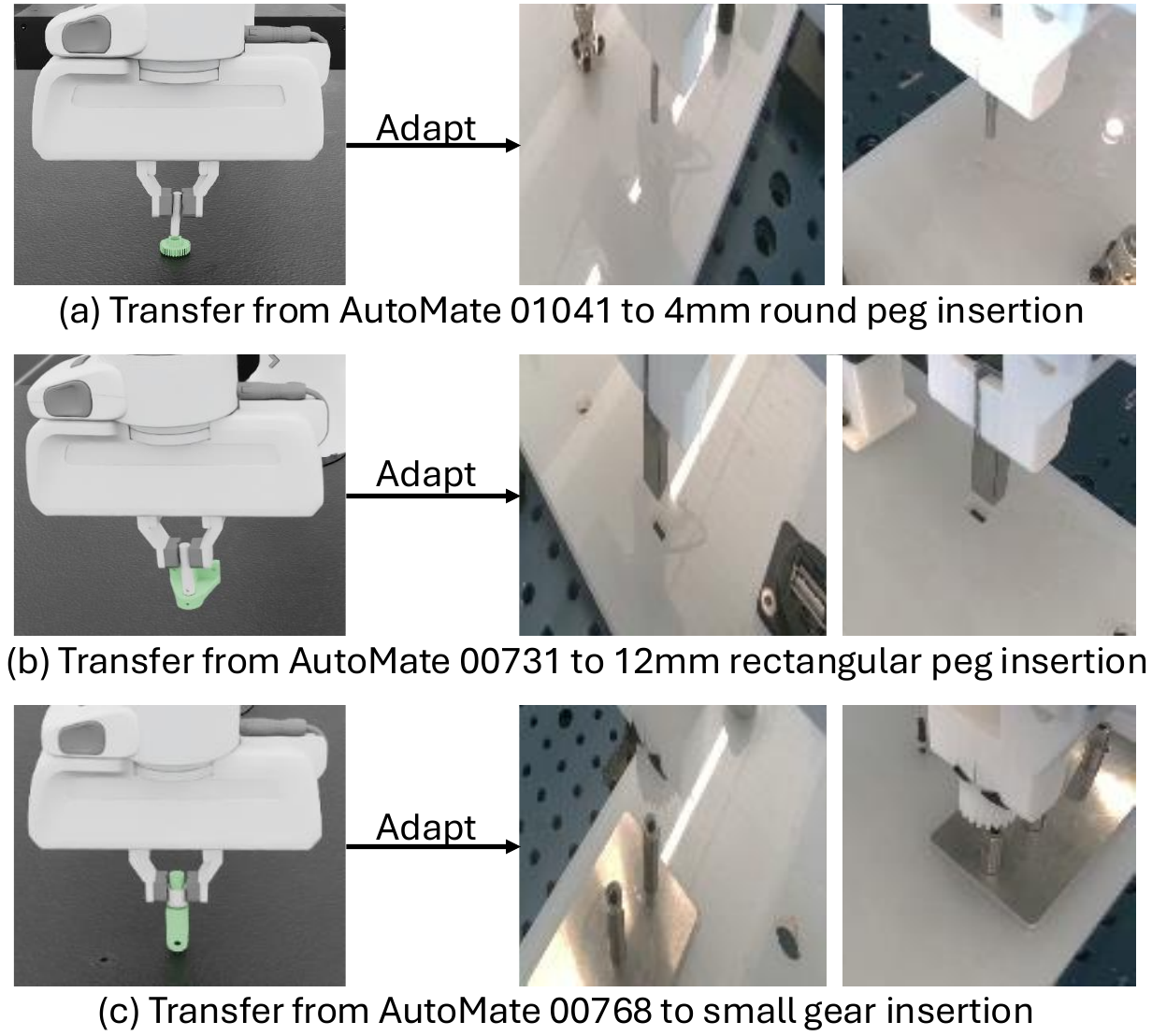}
  \vspace*{-0.06in}
  \caption{\small{\textbf{Adaptation of simulation policies from AutoMate tasks to NIST tasks.} We show images from wrist-mounted cameras here. Fig.~\ref{fig:task} (bottom) shows the NIST tasks from the front camera view.}}
  \label{fig:nist_adaptation}
  \vspace*{-0.3in}
\end{figure}

\vspace*{-0.05in}
\subsection{Generalization to Unseen Tasks}
\vspace*{-0.03in}

We evaluate the generalizability of {\ours} on NIST assemblies (Fig.~\ref{fig:task}) that were not seen during pre-training on AutoMate tasks. We aim to achieve strong real-world performance by adapting the base policy from relevant prior tasks.
In Fig.~\ref{fig:nist_adaptation}, we select the simulation policy based on the object size and the behavior pattern.
For \textit{$4mm$ round peg insertion}, we leverage AutoMate task 01041, which has the smallest plug ($6mm$ diameter) among the 10 AutoMate tasks.
For \textit{$12mm$ rectangular peg insertion}, we choose AutoMate task 00731, which has a $10mm$ diameter round peg.
For \textit{small gear insertion}, we select AutoMate task 00768, whose behavior involves placing a cap onto a cylinder, distinct from other typical peg-in-hole tasks. These choices result in reasonable zero-shot performance on the NIST tasks (success rates of 0.4–0.7). More systematic methods \cite{guo2025srsa} can be applied to reliably identify prior tasks transferring to new tasks according to task similarity and relevance.

When deploying the base policy, differences in dynamics between simulation and reality degrade the performance.
However, the policy is largely unaffected by differences in visual observations because it only conditions on low-dimensional state information.
As a result, the base policy still generates some successful demonstrations and serves as a functional prior for real-world training. We then train a residual policy using {\ours} to improve real-world success.

\begin{figure}[ht!]
\centering
    \includegraphics[width=0.95\linewidth]{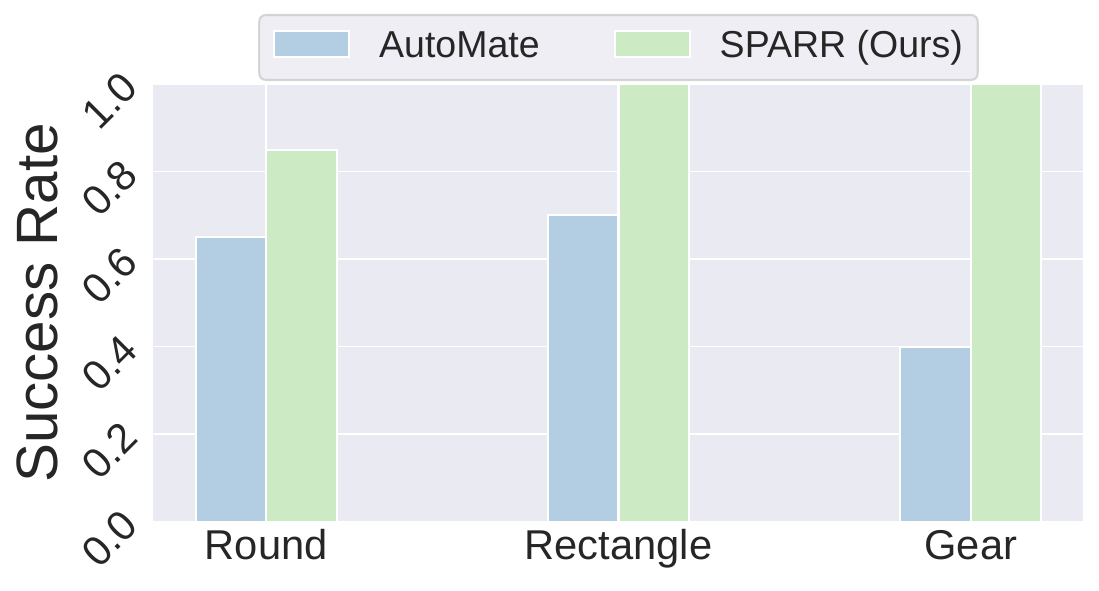}
\includegraphics[width=0.95\linewidth]{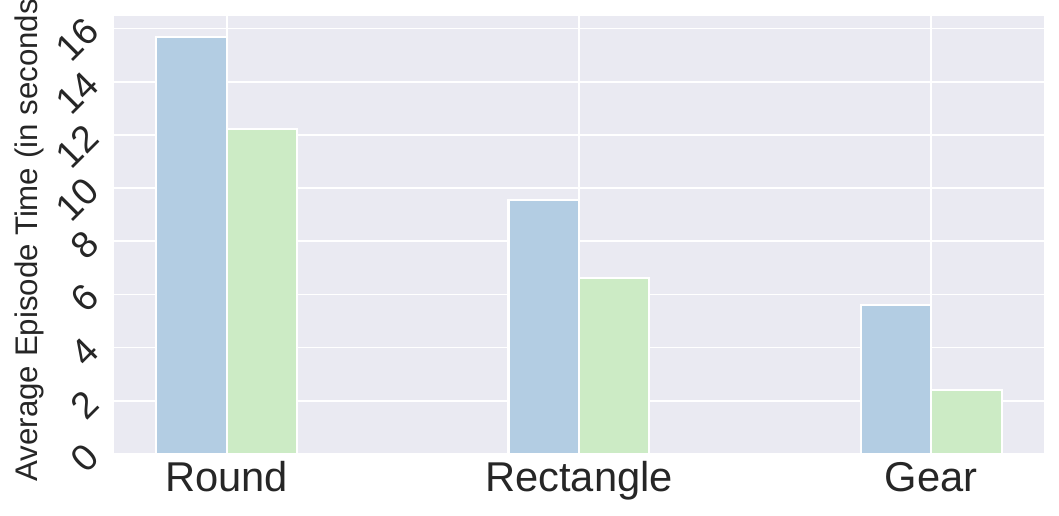}
    \vspace*{-0.1in}
    \caption{\small{\textbf{Performance on NIST assembly tasks.} {\ours} outperforms the baseline in success rate ($\uparrow$ higher is better) and cycle time ($\downarrow$ lower is better).
    }}
    \label{fig:nist_perf}
    \vspace*{-0.2in}
\end{figure}

\begin{figure}[ht!]
    \centering
    \includegraphics[width=0.98\linewidth]{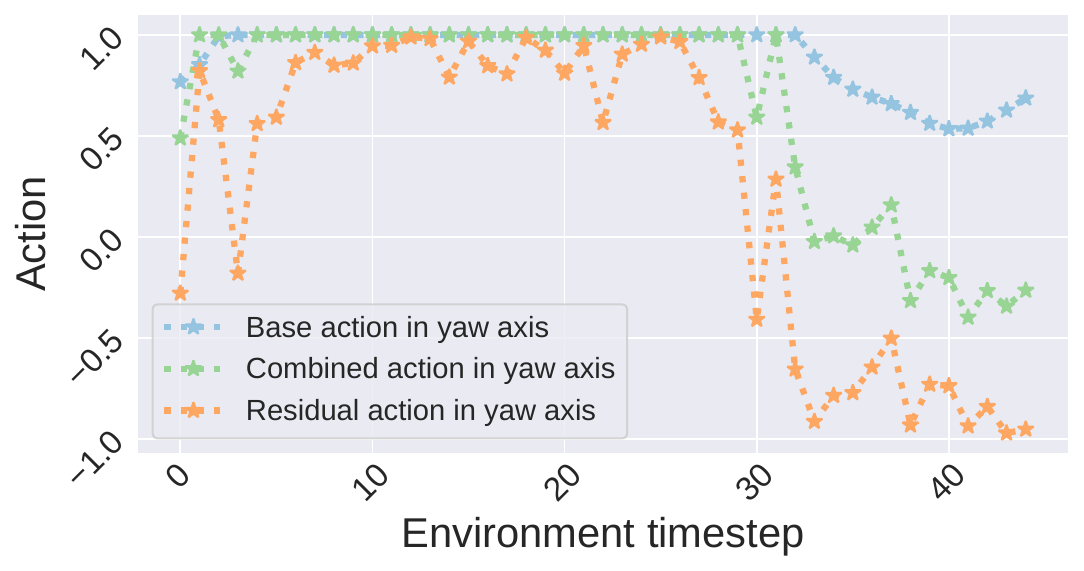}
    \vspace*{-0.1in}
    \caption{\small{\textbf{Visualization of base, residual and combined actions.} On a trajectory completing rectangular peg insertion task, the residual action in yaw axis disagrees with the base action and changes the plug rotation when the combined action is executed in the environment. All these actions are clamped to the range [-1,1] as incremental target pose.}}
    \label{fig:act_rect}
    \vspace*{-0.25in}
\end{figure}

Fig.~\ref{fig:nist_perf} shows the generalizability of {\ours} under unseen dynamics and visual observations.
For \textit{round peg insertion}, {\ours} achieves only an 80\% success rate because the tiny hole on the reflective white board is extremely difficult to detect (Fig.~\ref{fig:task}\&\ref{fig:nist_adaptation}). Although the goal is unclear in the image observation, the residual policy still helps to improve performance using input from the proprioceptive state.
For \textit{rectangular peg insertion}, the base policy was trained to insert a round peg and cannot handle the slight rotational mismatch with the rectangular hole. To emphasize this challenge, we add a uniform yaw noise of up to 3 degrees to the end-effector pose at the start of each episode.
Our residual policy learns the necessary rotation to align the peg with the target hole (Fig.~\ref{fig:act_rect}), achieving a perfect success in all 20 evaluation episodes.
For \textit{small gear insertion}, the gear clearance is much tighter ($0.005-0.014mm$) than the plug in AutoMate task 00768 ($0.5-1mm$). The residual policy learns to correct both the insertion angle and the applied actions, resulting in successful insertions. 
Overall, {\ours} achieves a relative improvement of 74.5\% in success rate and 36.5\% in cycle time across these NIST tasks.

\vspace*{-0.03in}
\subsection{Ablation Study}
\vspace*{-0.03in}
\label{sec:exp_ablation}

\paragraph{Effect of demo buffer update}
As described in Sec.~\ref{sec:training}, we update the demonstration buffer during RLPD policy learning. Specifically, we add high-quality trajectories that reach success faster than at least half of the offline demonstrations. This allows the policy to retain and exploit valuable experiences encountered during training, enabling it to complete the task faster. Table~\ref{tab:update_demo} compares methods with and without this demo buffer update, showing that {\ours} outperforms the variant that does not update the buffer.
\begin{table}[ht!]
\vspace*{-0.1in}
\caption{\small{\textbf{Comparison of method variants.} {\ours} shows higher success rate than the variant without demo buffer update.}}
\label{tab:update_demo}
\centering
\begin{tabular}{c|cccc}
\toprule
                & \multicolumn{2}{c}{Success rate ($\uparrow$)} & \multicolumn{2}{c}{Cycle time (s) ($\downarrow$)} \\
\toprule
Task            & 01041           & 01036          & 01041          & 01036         \\
\midrule
AutoMate            & 0.8             & 0.8            & 10.76          & 4.82          \\
\hline
{\ours} w/o demo update & 0.95            & 0.8            & 8.1            & \textbf{4.31}          \\
\hline
{\ours} (Ours) & \textbf{1.0}             & \textbf{1.0}            & \textbf{6.81}           & 5.45\\     
\bottomrule
\end{tabular}
\vspace*{-0.1in}
\end{table}

\paragraph{Effect of base action as input to residual policy}

Including the base action as an input provides important context for the residual correction. In Table~\ref{tab:base_input}, without the base action as input, the residual policy still slightly improves zero-shot deployment performance of AutoMate. compared to this variant without base action as input, {\ours} further enhances both the success rate and the cycle time.

\begin{table}[ht!]
\vspace*{-0.1in}
\caption{\small{\textbf{Comparison of method variants.} {\ours} outperforms the variant without base action as input to residual policy.}}
\label{tab:base_input}
\centering
\begin{tabular}{c|cccc}
\toprule
                & \multicolumn{2}{c}{Success rate ($\uparrow$)} & \multicolumn{2}{c}{Cycle time (s) ($\downarrow$)} \\
\toprule
Task            & 00015           & 00768          & 00015          & 00768        \\
\midrule
AutoMate            & 0.75             & 0.65            & 7.66          & 10.19          \\
\hline
{\ours} w/o base action input & 0.8            & 0.8            & 6.12            & 8.76         \\
\hline
{\ours} (Ours) & \textbf{1.0}             & \textbf{0.95}            & \textbf{3.51}           & \textbf{6.11}\\     
\bottomrule
\end{tabular}
\vspace*{-0.1in}
\end{table}

\vspace*{-0.05in}
\section{Conclusion and Future Work}
\vspace*{-0.03in}
In this work, we propose a residual policy learning approach {\ours} that leverages a simulation-trained state-based policy as a base and augments it with an asymmetric, vision-conditioned residual in the real world. The base policy provides structured priors and guides exploration, while the residual corrects for real-world discrepancies. On real-world two-part assembly tasks, {\ours} achieves near-perfect success without requiring human demonstrations or interventions, while also demonstrating robustness to state noise and generalizability to unseen tasks.

While {\ours} is highly effective, it relies on several assumptions. First, it requires the plug to be rigidly pre-grasped in the gripper. Investigating the full grasp-to-insertion pipeline and improving robustness to grasp perturbations remain topics for future work.
Second, {\ours} depends on reasonably good zero-shot sim-to-real transfer of the base policy. If the base policy achieves near-zero success in the real world, the residual component alone is insufficient to recover performance. Finally, {\ours} assumes access to automated reward or success detection during real-world deployment. Relaxing these assumptions is an important direction for future research.
It would be valuable to explore more expressive residual policies and more reliable success classifiers, such as multimodal models that integrate visual, force, and audio signals. Extending the framework to more diverse and challenging tasks beyond insertion-style assemblies is another promising direction.
\vspace*{-0.1in}

\bibliographystyle{IEEEtran}
\bibliography{IEEEabrv,references}

\end{document}